\documentclass[10pt,twocolumn,letterpaper]{article}
\usepackage[pagenumbers]{cvpr}
\usepackage{multirow}
\usepackage{array}
\usepackage{needspace}
\newcommand{\method}{\textit{MemoRenderer}}

\usepackage{algorithm}
\usepackage[noend]{algpseudocode}

\definecolor{linkblue}{rgb}{0.21,0.49,0.74}
\usepackage[pagebackref,breaklinks,colorlinks,allcolors=linkblue]{hyperref}
\hypersetup{pdfauthor={Ming Lin, Li Yin, Zehuan Huang, Yuzhen Wang, Yang-Tian Sun, Haoran Feng, Jianyi Zhang, Zhenghao Song, Hongxing Fan, Yan-Pei Cao, Lu Sheng},pdftitle={MemoRenderer: Code-Driven Generative World Rendering with Geometry-Aligned Memory}}
\title{MemoRenderer: Code-Driven Generative World Rendering with Geometry-Aligned Memory}
\author{
Ming Lin$^{1,*}$\quad Li Yin$^{1,*}$\quad Zehuan Huang$^{2}$\quad Yuzhen Wang$^{1}$\\[2pt]
Yang-Tian Sun$^{3}$\quad Haoran Feng$^{4}$\quad Jianyi Zhang$^{1}$\quad Zhenghao Song$^{1}$\\[2pt]
Hongxing Fan$^{5}$\quad Yan-Pei Cao$^{2}$\quad Lu Sheng$^{1}$\\[5pt]
{\small $^1$Beihang University\quad $^2$VAST\quad $^3$University of Hong Kong}\\
{\small $^4$Tsinghua University\quad $^5$Central University of Finance and Economics}\\[4pt]
{\small $^{*}$Equal contribution.\quad\url{https://dantargaryen.github.io/memorenderer-project-page/}}
}

\begin{document}

\twocolumn[
    \maketitle
    \vspace{-3em}
    \begin{center}
\centering
\includegraphics[width=\textwidth]{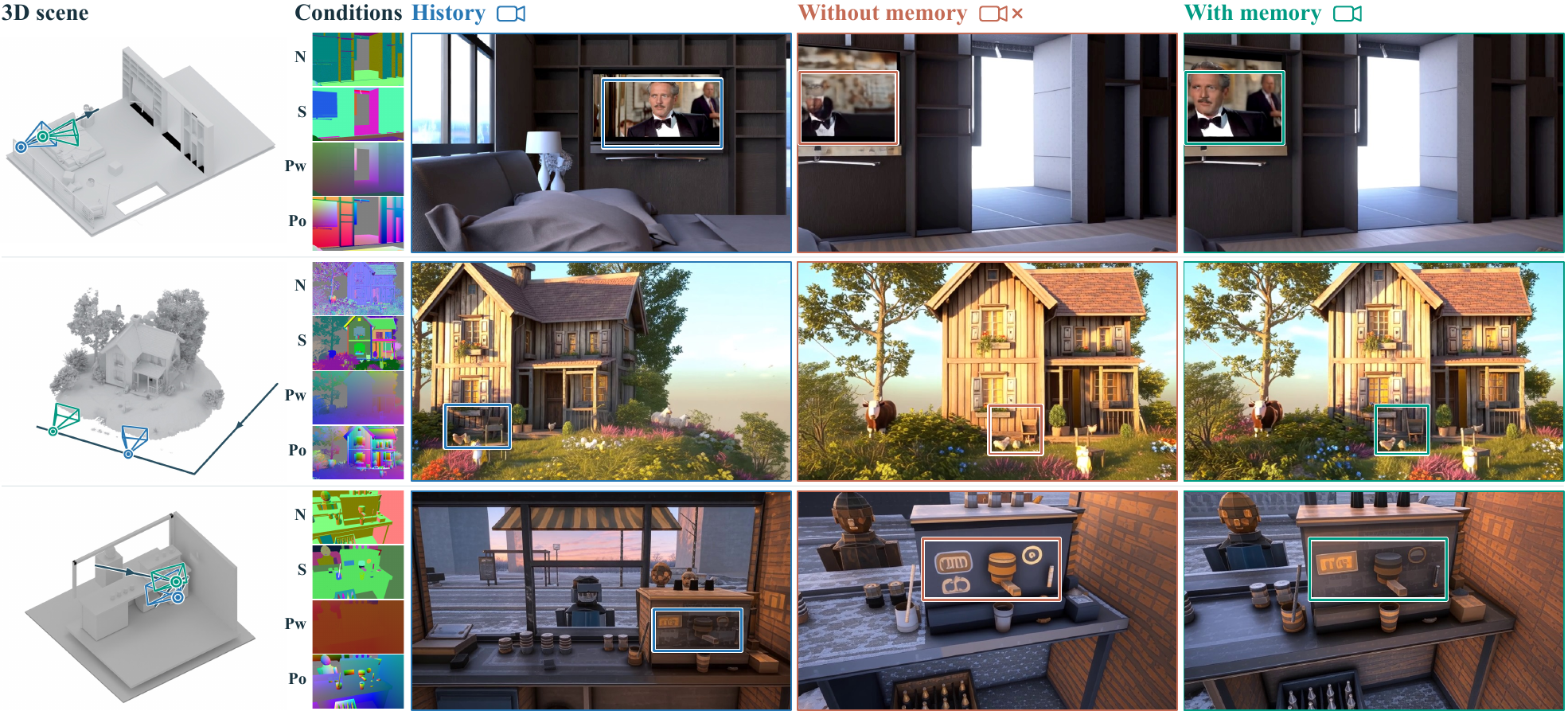}
\captionof{figure}{
\textbf{\method{} renders long-horizon memory-consistent videos in a streaming manner}, conditioned on engine-provided geometric maps, including normal maps, instance maps, world coordinate maps, and canonical coordinate maps, restoring appearance details like shadows across frames.
}
\label{fig:teaser}
\end{center}

    \bigbreak
]

\begin{abstract}
\begingroup
\hyphenpenalty=500
Code-driven world models use coding agents and engines to control world evolution, while video models synthesize appearance. However, persistent world state alone does not ensure consistent appearance beyond a streaming renderer's recent context.
We present \textit{MemoRenderer}, a generative world renderer with geometry-aligned visual memory.
Engine-provided normal and instance maps constrain scene structure, while world coordinate maps (WCMs) and instance-level canonical coordinate maps (CCMs) establish surface correspondences across viewpoints and object motion.
Code-driven retrieval uses these correspondences to select historical patches as memory context, achieving better appearance consistency than learnable memory baselines under the same conditioning.
To avoid searching an ever-growing history, a persistent, instance-based memory bank deduplicates historical observations and stores key-value entries from completed chunks.
Subsequent chunks search only this bank, reducing retrieval overhead during inference.
We introduce CodeMemBench to evaluate visual memory in static scenes and dynamic code-driven worlds.
Experiments show improved contextual appearance consistency over recent generative renderers.
\par
\endgroup
\end{abstract}

\section{Introduction}
\label{sec:intro}

Video world models aim to simulate how the state of an environment evolves in response to actions~\cite{bruce2024geniegenerativeinteractiveenvironments,sun2026worldplay,ha2018worldmodels,hafner2023dreamerv3,micheli2023iris,lin2026stateplay,meng2026marionette,zhu2026sanawm}. Many existing approaches pursue this goal through action-conditioned video generation, jointly modeling state evolution and visual synthesis~\cite{alonso2024diffusionworldmodelingvisual,valevski2025diffusionmodelsrealtimegame,hu2023gaia1,che2024gamegenx,zhang2025matrixgame,he2025matrixgame2,li2025hunyuangamecraft,huang2026solarwm}. More recently, code-driven world models have introduced an explicit separation between world execution and visual generation~\cite{chen2026cwm,huang2026pwm,yang2026oneiraopenendedgenerationopenworld,fang2026coder}. In this pipeline, a coding agent defines scene entities and interaction rules through code, while an engine executes the program, updates the world state, and renders proxy observations~\cite{chi2026agentgarten,zhan2026magpie}. A video renderer then uses these observations as conditioning inputs to synthesize the final imagery~\cite{vast2026edenlive,fang2026solarrenderer,lin2026gwrflash,ren2025gen3c,zhou2026proxyup,xu2026proxy2world}. This division enables explicit control over scene behavior and state transitions through code, while allowing the renderer to synthesize rich visual appearance from coarse engine outputs.

However, maintaining world state alone does not automatically ensure consistent visual appearance over time. Geometry and instance labels constrain what the renderer should depict, but much of the texture and fine detail is generated by the video model itself~\cite{cai2024generativerendering,liang2025diffusionrenderer,zeng2026rgbxnext}. These details should remain consistent as the camera moves, objects change pose, or previously occluded surfaces reappear~\cite{xiao2025worldmem,xu2026ucm,wang2026matrixgame3,xia2026loci}. As interaction continues, new observations also reveal surfaces absent from the initial frame, expanding the visual history that must be retained. Yet a renderer limited to recent frames may lose access to this history~\cite{yu2025contextmemory,yi2026worldkv,guo2025long,cai2025mixture,li2025stable,streamingt2v,zhang2026worldplay2}, even when the engine continues to track the same entities. Visual memory is therefore a distinct requirement at the rendering interface.

Prior work in traditional action-conditioned world models has explored two broad approaches to access memory context. One retrieves relevant past observations using predefined rules based on view overlap or reconstructed surfaces~\cite{yu2025contextmemory,li2025vmem,qian2026matrixgame35,liao2026memorizon}. The other adaptively extracts and integrates historical information through learned context queries, attention routing, or test-time updates~\cite{yu2026memlearner,xu2026sparseforcingnativetrainable,zhang2025testtimetrainingright,wu2026worldt3,xiao2025worldmem,xia2026loci,dalal2025one}. In code-driven rendering, the engine directly provides 3D surface coordinates, instance identities, and tracking information, offering a direct basis for associating the current view with past observations. These cues can serve as conditioning inputs that help the video model identify surface correspondences over time. They can also guide memory organization and the retrieval of relevant historical appearance. This leads us to two research questions:
\begin{enumerate}[leftmargin=2em]
    \item \emph{What should engine-provided geometry be represented as for video conditioning so that the model can effectively use spatial and temporal correspondences?}
    \item \emph{Given reliable geometry and tracking, are explicit geometric retrieval or learned memory networks more effective at generating consistent appearance?}
\end{enumerate}

To address these issues, we introduce \method{}, a generative world renderer that achieves consistent geometry-aligned visual memory.
For geometric conditioning, we use normal and instance maps to introduce structural and instance region controls.
Additionally, world coordinate maps (WCMs) are used to link different observations of the same static scenes, and canonical coordinate maps of instances (CCMs) are introduced to establish temporal correspondences of moving objects.
This geometric representation provides a strong positional encoding for video diffusion models to learn geometry-aligned consistency.

For memory modeling, we build a code-driven retrieval baseline.
When generating one chunk, the system searches for relevant historical patches based on WCMs, CCMs and instance maps, and selects them as memory contexts.
This code-driven baseline is superior to learnable memory approaches in maintaining consistent appearance.
However, it retrieves relevant patches across all histories, resulting in a redundant search space.
Therefore, we further introduce a persistent memory bank anchored to the finite 3D scene and organized by instance.
After each video chunk, we update the bank with its key-value tokens, using instance maps to associate tokens with instances and CCMs to deduplicate the pre-observed surface regions.
The bank thus grows with surface coverage rather than video duration, remaining bounded and compact, allowing the model to learn clearer associations with historical context.

We construct CodeMemBench to evaluate visual memory in static scenes and dynamic code-driven worlds, including camera motion, occlusion, moving instances. Experimental results show that \method{} improves contextual appearance consistency over recent generative renderers, including a MiniMax-H3-based renderer~\cite{fang2026solarrenderer}.
Our contributions are summarized as follows:
\begin{itemize}[leftmargin=1.5em]
    \item Geometric conditioning that combines normal and instance maps with WCMs and CCMs to expose surface correspondences across time.
    \item Code-driven retrieval with a persistent, instance-based deduplicated memory bank, achieving better appearance consistency than learnable memory baselines.
    \item CodeMemBench and a history-based evaluation protocol for studying visual quality, structural alignment, and appearance consistency in interactive worlds.
\end{itemize}

\section{Related Work}
\label{sec:related_work}

\paragraph{Code World Models.}
A recent trend in interactive world modeling is to decouple state evolution from visual rendering~\cite{lin2026gamedirector}, using programs and engines to maintain world state evolution, and video models as renderers~\cite{chen2026cwm,huang2026pwm,huang2026generativeworldrenderer,lin2026gwrflash}. 
The Project Eden preview describes a shared-state interface between world evolution and generative rendering~\cite{vast2026projecteden}.
Code World Model, Programmable World Model, Oneira, and CoDeR use coding agents and executable programs to maintain explicit world states~\cite{chen2026cwm,huang2026pwm,yang2026oneiraopenendedgenerationopenworld,fang2026coder}.
AgentGarten and Magpie connect simulators or game engines to generative rendering~\cite{chi2026agentgarten,zhan2026magpie}, while EdenLive and SolarRenderer provide real-time rendering interfaces from any visual reference~\cite{vast2026edenlive,fang2026solarrenderer}.
\method{} follows this decoupled paradigm and focuses on visual consistency in rendered long-term video, investigating how engine-provided surface coordinates and instance associations can support \mbox{long-term visual consistency}.

\paragraph{Geometry-Conditioned Video Generation.}
Geometry-conditioned video models use depth maps~\cite{2023videocomposer,guo2024sparsectrl,alhaija2025cosmos,gao2025longvie}, point tracks~\cite{wang2023motionctrl,shi2024motion}, or rendering buffers for structural and motion control~\cite{vace,gao2025longvie2,gu2025das,huang2026generativeworldrenderer,he2024cameractrl,burgert2025go}. 
 Other methods accept sparse 3D layouts or coarse rendered proxies~\cite{bhat2026loosecontrolvideodirectorialvideocontrol,gomeznogales2026coarsetoreal,odin2026_3dreal,cai2024generativerendering,ren2025gen3c,zhou2026proxyup,xu2026proxy2world,gu2026gorenderer}. 
 VideoFrom3D combines geometry-derived edges and optical flow~\cite{kim2025videofrom3d}, 
 while Totem uses canonical surface anchors to preserve object consistency~\cite{wan2026totem}. 
 \method{} conditions video generation on normal maps, instance maps, WCMs, and CCMs, providing a strong positional encoding for learning geometry-aligned visual consistency. These engine-provided correspondences also support the retrieval of historical appearance. 

\paragraph{Memory for Video Generation.}
Visual memory for long video generation is commonly built through rule-based retrieval~\cite{yu2025contextmemory,liang2025lookingbackward,wang2026matrixgame3,liao2026memorizon} or learnable memory modeling~\cite{yu2026memlearner,xiao2025worldmem,xu2026ucm,xia2026loci,streamingt2v,guo2025long,cai2025mixture,zhu2026sanawm,zhang2026worldplay2}. 
 Context as Memory and WorldPlay retrieve past frames using view overlap and add them to the context~\cite{yu2025contextmemory,sun2026worldplay}. MosaicMem and Matrix-Game 3.5 retrieve geometrically aligned patches~\cite{yu2026mosaicmem,qian2026matrixgame35}, while WorldKV reuses stored KV within the native attention window without re-encoding~\cite{yi2026worldkv}. Learnable methods query history through adaptive tokens or sparse attention~\cite{yu2026memlearner,xu2026sparseforcingnativetrainable,cai2025mixture}, or consolidate observations into fast-weight states through test-time updates~\cite{zhang2025testtimetrainingright,wu2026worldt3,dalal2025one}. In code-driven world models, we find that retrieval based on engine-provided geometric correspondences maintains appearance consistency more effectively than learnable memory mechanisms. We further introduce a persistent, instance-based memory bank to reduce redundant observations and improve retrieval efficiency. 

\section{Method}
\label{sec:method}

\subsection{Overview}
\label{subsec:method_overview}

\method{} renders detailed RGB video from engine-provided condition maps, an initial image, and a text prompt. The engine maintains scene state and geometric correspondence, while the video model synthesizes appearance and retains its visual consistency. We connect these roles through a shared geometric description that guides both current conditioning and access to historical patch KV.

Let $I_0$ denote the initial RGB frame, $p$ the text prompt, and $\mathbf C_i$ the condition latents for chunk $i$. Given the current conditions and available history, the renderer generates a clean latent chunk autoregressively:
\begin{equation}
\widehat{\mathbf Z}_i
=
G_\theta\!\left(
\boldsymbol\epsilon_i;
\mathbf C_i,I_0,p,\mathcal H_i
\right),
\label{eq:method_chunk_generation}
\end{equation}
Here, $\boldsymbol\epsilon_i$ is the initial noise, and $G_\theta$ denotes the few-step generator. The context $\mathcal H_i$ combines recent observations with the retrieved historical patch KV cache. Once a chunk is finalized, its clean-context features and geometric metadata become available to later predictions.

\begin{figure}[t]
  \centering
  \includegraphics[width=\linewidth]{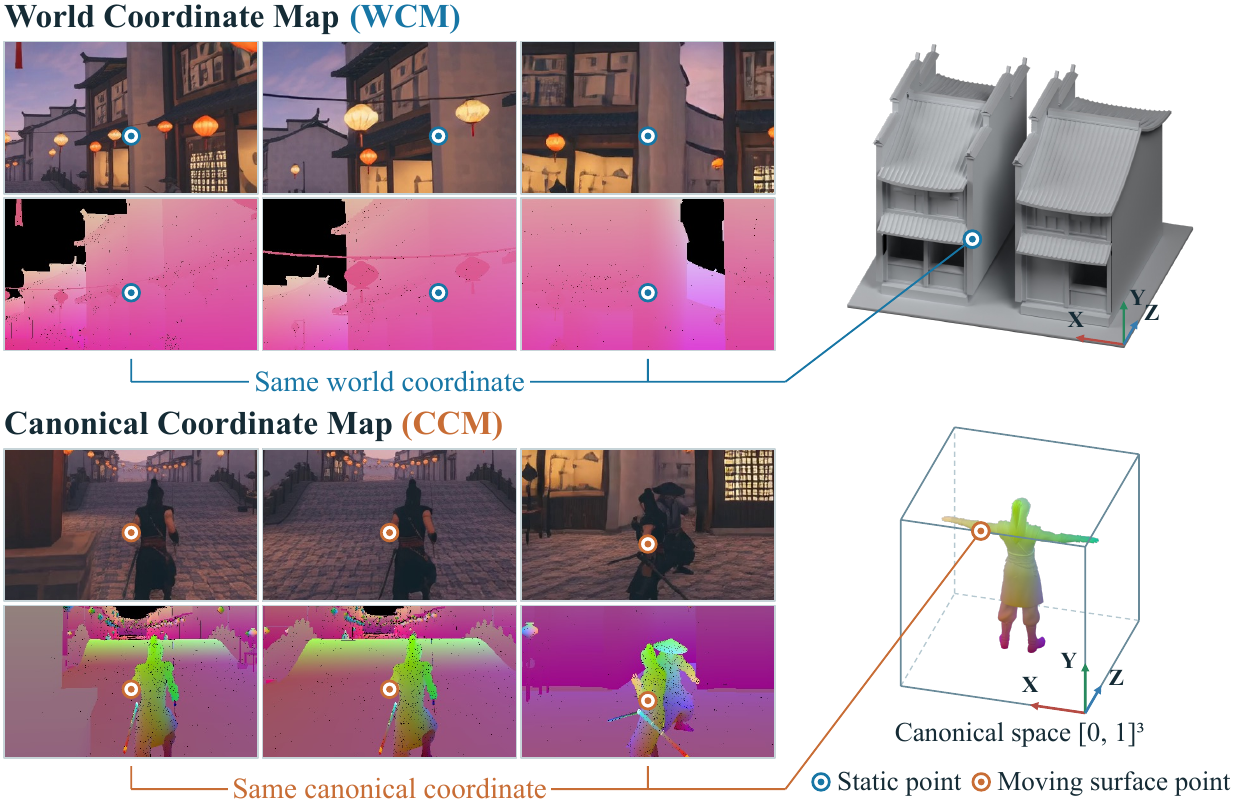}
  \caption{\textbf{World and canonical coordinate map overview.} WCMs share the same coordinates for static scenes across different frames, while CCMs provide tracking for dynamic instances.}
  \label{fig:condition_map}
\end{figure}

\begin{figure*}[t]
  \centering
  \includegraphics[width=\linewidth]{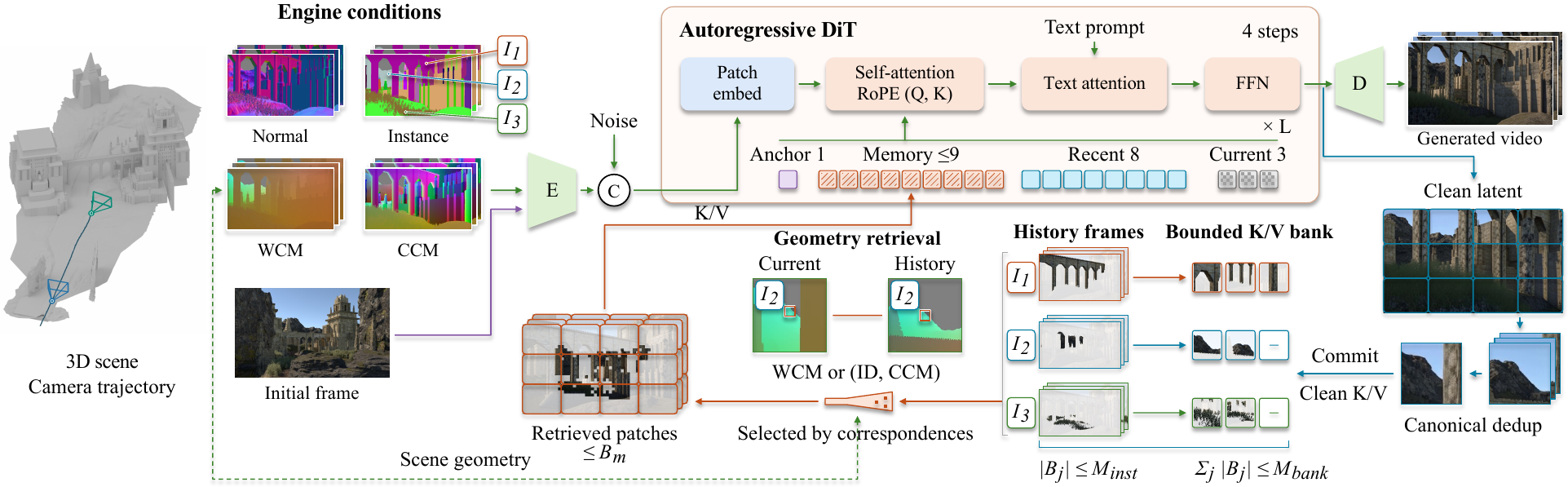}
  \caption{\textbf{Pipeline overview.} Engine-provided geometry guides scene structure and motion, while historical visual features, retrieved from a persistent memory bank, support appearance consistency across viewpoint changes and object motion.}
  \label{fig:framework}
\end{figure*}

\subsection{Geometric Conditioning}
\label{subsec:geometry_memory_conditioning}

We condition the video model on four engine-rendered geometric maps: normal maps, instance maps, world coordinate maps (WCMs), and instance-level canonical coordinate maps (CCMs).

Normal maps $N$ record the world-space surface normal at each visible 3D surface point, providing local geometric structure. Instance maps $S$ assign each scene instance a unique ID and a distinct RGB color, partitioning instance regions within each frame while keeping the same color across frames to provide temporal instance guidance. WCMs $P^w$ record the 3D world-space coordinates of visible surface points. Since static assets dominate most scenes, the same spatial point retains an identical WCM pixel value across frames as the camera moves, providing a direct mapping for modeling static-scene appearance consistency. CCMs $P^o$ normalize each instance asset into a fixed, bounded canonical coordinate system. For a visible pixel $u$ of an instance, $P^o(u) \in \mathbb{R}^3$ stores the canonical coordinate of the observed surface point. Because this canonical space remains invariant regardless of how the asset moves, the same surface point on a dynamic object maintains a fixed pixel value across frames, establishing temporal correspondences for moving instances (\cref{fig:condition_map}).

After rendering these four geometric condition videos with the engine, we encode each one with a frozen VAE~\cite{kingma2013auto,rombach2022high,wan2025}. The resulting condition latents $\mathbf C_i$ are concatenated with the noisy RGB latents of the current video chunk along the channel dimension, and patch embedding converts the concatenated latents into video DiT~\cite{peebles2023scalable} tokens, providing positional encoding for geometry-aligned generation.

\subsection{Few-Step AR Distillation}
\label{subsec:long_horizon_training}

To enable real-time autoregressive rendering, we distill a pretrained video diffusion model into a few-step causal generator~\cite{yin2024causvid,huang2025self,cui2025selfforcingpp,zhu2026causalforcing,zhao2026causalforcingpp,zheng2026causalrcm,zhao2026minwm}. We first perform supervised fine-tuning (SFT) on the pretrained bidirectional model with the four geometric condition maps, then adapt it to causal chunk-by-chunk generation via teacher-forcing (TF) SFT, and finally distill it with Self Gradient Forcing (SGF)~\cite{zhuang2026sgf} for few-step autoregressive generation.

During distillation, the student performs autoregressive rollouts and aligns its output distribution with the geometry-conditioned bidirectional teacher via distribution matching distillation (DMD)~\cite{yin2024one,yin2024improved}, where $\mathcal{L}_{\mathrm{DMD}}$ matches the student and teacher distributions using real- and fake-score networks. To further preserve perceptual fidelity, we combine $\mathcal{L}_{\mathrm{DMD}}$ with a perceptual LPIPS~\cite{zhang2018unreasonable} loss $\mathcal{L}_{\mathrm{LPIPS}}$ between decoded student frames and paired ground-truth RGB:
\begin{equation}
\mathcal{L}_{\mathrm{distill}} = \mathcal{L}_{\mathrm{DMD}} + \lambda_{\mathrm{LPIPS}} \mathcal{L}_{\mathrm{LPIPS}}.
\label{eq:method_distill_loss}
\end{equation}
Following SGF, each training step first runs a gradient-free autoregressive rollout to record the generated trajectory, and then reconstructs the supervised prediction on this recorded trajectory so that gradients flow through the recomputed historical context encoding and current chunk synthesis.

In vanilla streaming inference~\cite{chen2025diffusion} without dedicated memory context, the renderer attends only to a fixed sliding window~\cite{gao2025ca2,yang2025longlive} of $1$ initial sink latent frame and $17$ recent latent frames alongside the current $3$-frame chunk ($21$ latent frames in total). Because earlier observations fall out of this local window as the rollout progresses, this sliding-window scheme struggles to maintain visual consistency with previously generated history when revisiting earlier regions.

\begin{figure*}[t]
  \centering
  \includegraphics[width=\linewidth]{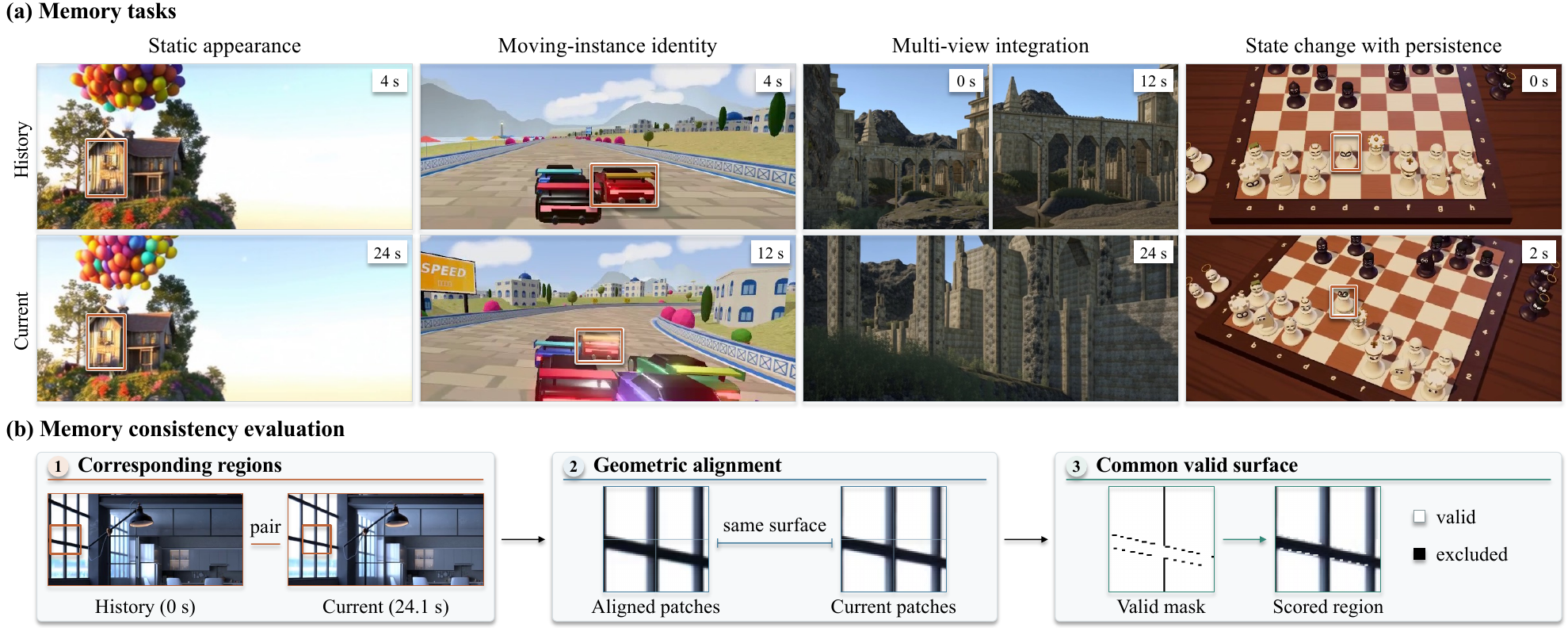}
  \caption{\textbf{\textit{CodeMemBench} and memory consistency evaluation.} (a) \textit{CodeMemBench} includes comprehensive evaluation on static appearance persistence, moving-instance identity, multi-view integration, and program-controlled state changes. (b) The benchmark computes alignment metrics on geometrically aligned historical and current patches for memory evaluation.}
  \label{fig:codemembench}
\end{figure*}

\subsection{Code-Driven Memory Retrieval}
\label{subsec:historical_patch_kv}


\begin{algorithm}[t]
\caption{Code-Driven Memory Retrieval and Update}
\label{alg:memory_retrieval}
\footnotesize

\newcommand{\phase}[1]{%
    \Statex \vspace{0.35em}\textbf{#1}}
\newcommand{\algNote}[1]{%
    \Statex {\scriptsize\itshape
    \(\triangleright\)~#1}}

\begin{algorithmic}[1]
\Require Geometry $Q_i$, bank $\mathcal B_i$,
budgets $B_m, M_{\mathrm{inst}}, M_{\mathrm{bank}}$

\phase{1. Retrieve memory}
\For{each visible query patch $q \in Q_i$}
    \State $\mathcal L_q \gets \emptyset$
    \State $j \gets \textsc{InstanceID}(q)$
    \If{$j \in \mathcal B_i$}
        \State $\mathcal C_q \gets
        \textsc{Lookup}(\mathcal B_i[j], \mathrm{CCM}_q)$
        \algNote{Check instance, WCM, and projected footprint}
        \State $\mathcal L_q \gets
        \textsc{Verify}(\mathcal C_q, q)$
    \EndIf
\EndFor
\algNote{Rank, round-robin select, and deduplicate within $B_m$}
\State $(K_m,V_m) \gets
\textsc{RetrieveKV}(\mathcal B_i,\{\mathcal L_q\},B_m)$

\phase{2. Generate chunk}
\algNote{Denoise with anchor, retrieved memory, and recent context}
\State $z \gets
\textsc{Generate}(Q_i,[K_a;K_m;K_r],[V_a;V_m;V_r])$

\phase{3. Update memory}
\algNote{Extract clean KV and insert by instance and canonical cell}
\State $(K,V) \gets \textsc{CleanForward}(z)$
\State $\mathcal B_{i+1} \gets
\textsc{Insert}(\mathcal B_i,K,V,Q_i)$
\algNote{Retain representative views only when capacity is exceeded}
\State $\mathcal B_{i+1} \gets
\textsc{Prune}(\mathcal B_{i+1},M_{\mathrm{inst}},M_{\mathrm{bank}})$
\end{algorithmic}
\end{algorithm}

To mitigate long-term appearance drift, we use the same geometry-based retrieval rules during distillation and inference. We retrieve historical clean KV before denoising each chunk and update an instance-organized memory bank after the chunk is generated (\cref{alg:memory_retrieval}). For each visible query patch, instance labels and CCM coordinates identify historical observations of the same object surface, while WCM and projected-footprint checks reject mismatches. Within the retrieval budget, we prioritize the earliest matches, followed by coverage, density, and view alignment, using round-robin selection and uniform filling.

As shown in \cref{fig:framework}, each attention layer combines
anchor, retrieved-memory, recent-history, and current-chunk KV
under a fixed context budget.
Historical keys are cached before RoPE~\cite{su_roformer}.
When assembling the context, we preserve their spatial indices
and apply RoPE once using compact temporal indices within
a 21-latent-frame span.

To reduce redundant observations and bound memory growth,
we organize the bank by instance and canonical-grid cell.
After each chunk is finalized, a clean forward pass extracts
its KV, which are stored with source indices and geometry.
When capacity is exceeded, we retain representative views
subject to $|\mathcal B_i[j]|\le M_{\mathrm{inst}}$ and
$\sum_j|\mathcal B_i[j]|\le M_{\mathrm{bank}}$,
prioritizing earliest observations, followed by dispersed
recent views and uniform filling.
Further details are provided in the supplementary material.

Under the same framework, geometric retrieval better preserves
appearance consistency than learned-router retrieval,
sparse-attention retrieval, and TTT-based fast-weight memory
(\cref{sec:experiment}).


%
%

\section{Experiments}
\label{sec:experiment}

\subsection{Implementation Details}
\label{subsec:implementation}
\label{subsec:experiment_data}

\paragraph{Training Dataset.}
We construct a paired rendering corpus from diverse open-source Unreal Engine and Blender environments. It comprises 20,000 81-frame clips spanning dozens of scenes, split by source session into 19,000 training and 1,000 validation clips. Each clip pairs RGB frames with normal maps, instance maps, WCMs, and CCMs at $832\times480$ and 16 fps, alongside numerical geometry and camera metadata for cross-frame surface correspondences. From the same scene pool, we further extract 4,000 longer sequences (161, 241, 321, or 397 frames) under the same session split, excluding duplicate revisit events, to expose the renderer to history beyond its recent context.

\begin{table}[t]
  \centering
  \caption{\textbf{Quantitative comparison on 81-frame sequences.}}
  \label{tab:renderer_comparison}
  \footnotesize
  \setlength{\tabcolsep}{1pt}
  \renewcommand{\arraystretch}{0.95}
  \begin{tabular*}{\columnwidth}{@{\extracolsep{\fill}}lrrrrrr@{}}
    \toprule
    \multirow{2}{*}{Method} & \multicolumn{2}{c}{\shortstack{Visual\\Quality}} & \multicolumn{2}{c}{\shortstack{Visual\\Alignment}} & \multicolumn{2}{c}{\shortstack{Structure\\Alignment}} \\
    \cmidrule(lr){2-3}\cmidrule(lr){4-5}\cmidrule(lr){6-7}
      & \shortstack{MUSIQ\\$\uparrow$} & \shortstack{DOVER\\$\uparrow$} & \shortstack{PSNR\\$\uparrow$} & \shortstack{SSIM\\$\uparrow$} & \shortstack{PSNR-D\\$\uparrow$} & \shortstack{Ctrl.\\DINO$\uparrow$} \\
    \midrule
    VACE~\cite{vace} & 64.6825 & 0.5330 & 12.2607 & 0.4121 & \textbf{20.6625} & 0.6965 \\
    DaS~\cite{gu2025das} & 55.3964 & 0.4970 & 16.5021 & 0.5440 & 18.8653 & 0.6477 \\
    LongVie 2~\cite{gao2025longvie2} & 58.3286 & 0.5441 & 19.9844 & 0.6748 & 20.4677 & 0.7554 \\
    VideoFrom3D~\cite{kim2025videofrom3d} & 60.0605 & \underline{0.5839} & 12.5876 & 0.4712 & 19.5908 & 0.6298 \\
    C2R~\cite{gomeznogales2026coarsetoreal} & 65.8548 & 0.5703 & 12.4536 & 0.4511 & 19.9913 & 0.6433 \\
    LTX-2.3~\cite{odin2026_3dreal} & 55.5820 & 0.4519 & 22.2527 & 0.7210 & 20.2622 & 0.7390 \\
    AgentGarten~\cite{chi2026agentgarten} & \underline{65.9587} & 0.5621 & 21.4913 & 0.6851 & 20.2258 & 0.7522 \\
    Streaming CWM~\cite{liao2026streamingcwm} & 61.2230 & 0.5029 & 18.5773 & 0.5988 & 19.7875 & 0.7194 \\
    SolarRenderer~\cite{fang2026solarrenderer} & 62.3441 & 0.5403 & \underline{23.3945} & \underline{0.7579} & 20.5057 & \underline{0.7669} \\
    EdenLive-5B~\cite{vast2026edenlive} & 56.8473 & 0.4763 & 19.3588 & 0.6264 & 20.1544 & 0.7458 \\
    EdenLive-14B~\cite{vast2026edenlive} & 63.9326 & 0.5536 & 21.1963 & 0.6712 & 20.1465 & 0.7452 \\
    \textbf{MemoRenderer} & \textbf{66.8629} & \textbf{0.5846} & \textbf{25.4171} & \textbf{0.8151} & \underline{20.5092} & \textbf{0.7698} \\
    \bottomrule
  \end{tabular*}
\end{table}

\begin{figure*}[t]
  \centering
  \includegraphics[width=\linewidth]{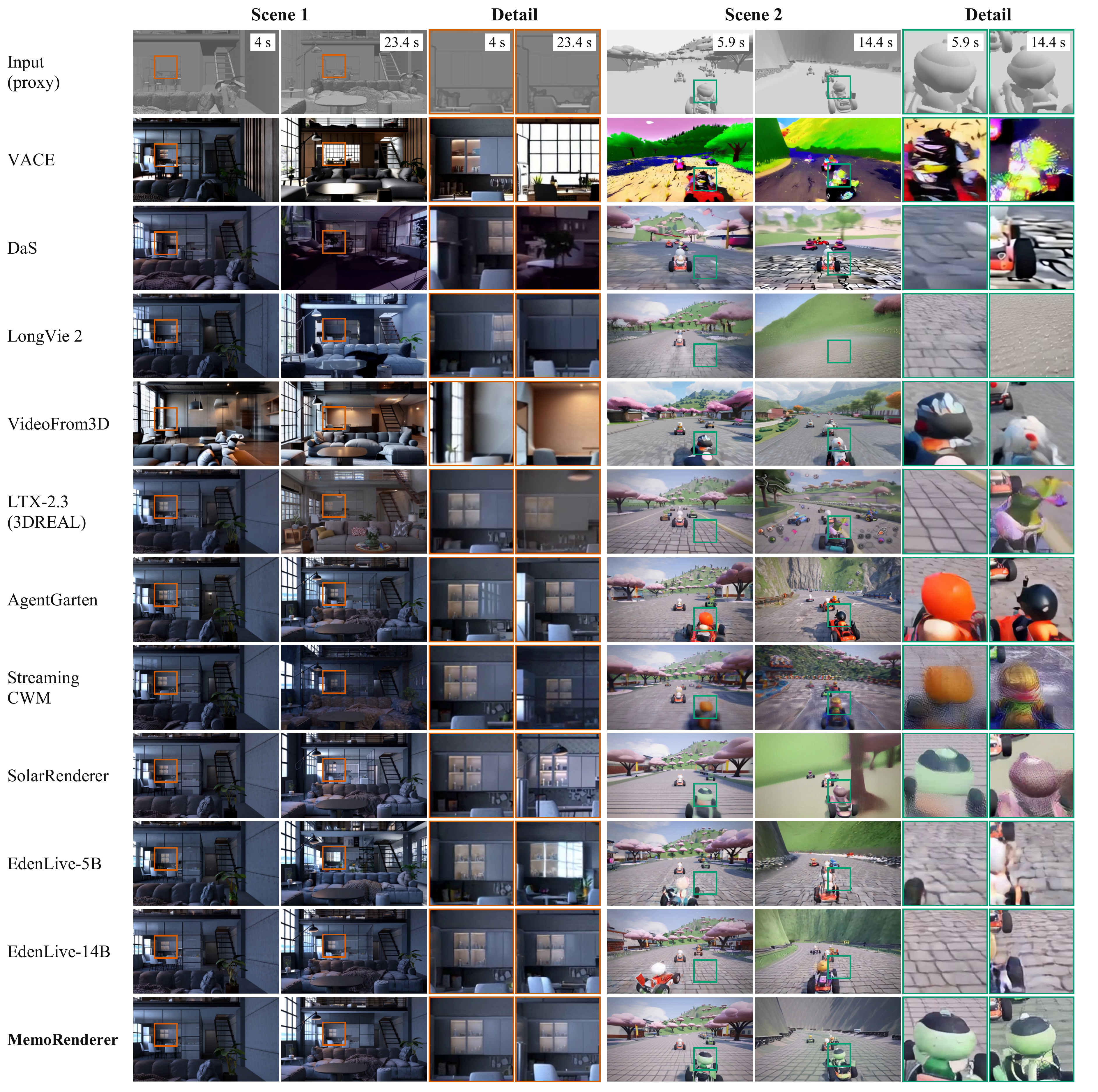}
  \caption{\textbf{Qualitative comparisons on a static interior scene and a dynamic kart-racing scene.} Each scene shows two generated observations and paired enlargements of the highlighted corresponding regions.}
  \label{fig:comparison_short_term}
\end{figure*}

\paragraph{Training Details.}
We build our renderer on the Wan2.1-1.3B~\cite{wan2025} backbone, initializing the generator from the teacher-forcing model and the real- and fake-score networks from the bidirectional model. Following \cref{subsec:long_horizon_training}, we distill the base renderer on 81-frame clips using Self Gradient Forcing~\cite{zhuang2026sgf} and retain exponential-moving-average generator weights for evaluation. Prior to quantitative evaluation, MemoRenderer is further fine-tuned on long-memory episodes. To ensure fair comparisons, all conditioning and memory ablation variants follow the same backbone, training pipeline, and evaluation protocol, differing only in the ablated inputs or memory strategy.

For long-memory adaptation, we conduct a separate training stage using the 4,000-sequence corpus, with 3,800 sequences for training and 200 for validation. We train on variable-length revisit episodes that retain the generated history from the sequence start to a query chunk near the revisit event. Supervision is applied only to the query chunk.

\paragraph{Inference Details.}
Inference uses four denoising steps, classifier-free guidance scale 1, deterministic sampling, and clean-history KV. Each regular chunk contains three latent frames. The active context contains at most 21 latent-frame equivalents, including the current chunk. Geometry-guided retrieval allocates one frame to the initial anchor, up to nine frame-equivalents to historical patch KV, and eight frames to recent history. Temporal RoPE indices are reassigned within this context. The memory strategies share the same backbone architecture, initial inputs, random seed, sampling schedule, and context limit. Each strategy uses its corresponding trained weights, fixed during inference.

\subsection{Evaluation Benchmark}
\label{subsec:evaluation_benchmark}
\label{subsec:experiment_baselines}
\label{subsec:experiment_metrics}

\begin{table*}[tp]
  \centering
  \caption{\textbf{Quantitative comparison and ablation on long-clip rendering}, including (A) comparison with baselines, ablation studies on (B) geometric conditioning, and (C) memory access strategies.}
  \label{tab:long_rollout}
  \label{tab:renderer_comparison_long}
  \label{tab:dynamic_scene_comparison}
  \label{tab:conditioning_ablation}
  \label{tab:memory_ablations}
  \label{tab:memory_mechanisms}
  \footnotesize
  \setlength{\tabcolsep}{3pt}
  \renewcommand{\arraystretch}{0.95}
  \begin{tabular*}{\textwidth}{@{\extracolsep{\fill}}lrrrrrcrrr@{}}
    \toprule
    \multirow{2}{*}{Method / Setting} & \multicolumn{2}{c}{Visual Quality} & \multicolumn{2}{c}{Visual Alignment} & \multicolumn{2}{c}{Structure Alignment} & \multicolumn{3}{c}{Memory Consistency} \\
    \cmidrule(lr){2-3}\cmidrule(lr){4-5}\cmidrule(lr){6-7}\cmidrule(lr){8-10}
      & MUSIQ$\uparrow$ & DOVER$\uparrow$ & PSNR$\uparrow$ & SSIM$\uparrow$ & PSNR-D$\uparrow$ & Ctrl. DINO$\uparrow$ & M.MSE$\downarrow$ & M.PSNR$\uparrow$ & M.DINO$\uparrow$ \\
    \midrule
    \multicolumn{10}{l}{\textit{(A) Overall comparison (50\% static + 50\% dynamic)}} \\
    \addlinespace[2pt]
    VACE~\cite{vace} & 61.7319 & 0.4639 & 9.7463 & 0.3602 & \underline{23.3528} & 0.6502 & 0.065072 & 13.7348 & 0.5385 \\
    DaS~\cite{gu2025das} & 50.0763 & 0.3892 & 11.5544 & 0.4285 & 21.4973 & 0.4622 & 0.076807 & 12.6263 & 0.3730 \\
    LongVie 2~\cite{gao2025longvie2} & 60.0026 & 0.4816 & 13.4697 & 0.4801 & 23.0324 & 0.6283 & 0.061780 & 13.8994 & 0.5224 \\
    VideoFrom3D~\cite{kim2025videofrom3d} & 60.8394 & \textbf{0.5469} & 12.3375 & 0.4788 & 22.6472 & 0.5820 & 0.057683 & 13.9323 & 0.5087 \\
    LTX-2.3 (3DREAL)~\cite{odin2026_3dreal} & 54.7585 & 0.4294 & 14.8994 & 0.5468 & 22.6620 & 0.5869 & 0.045296 & 13.9644 & 0.4792 \\
    AgentGarten~\cite{chi2026agentgarten} & \underline{66.6623} & 0.4911 & 15.8837 & 0.5648 & 23.0618 & 0.6704 & \underline{0.021994} & \underline{18.0980} & \underline{0.6097} \\
    Streaming CWM~\cite{liao2026streamingcwm} & 60.9356 & 0.4390 & 14.1782 & 0.4756 & 22.9055 & 0.6118 & 0.035005 & 15.4675 & 0.5113 \\
    SolarRenderer~\cite{fang2026solarrenderer} & 62.3944 & 0.4583 & 14.3783 & 0.5580 & 23.1427 & \underline{0.6937} & 0.051729 & 13.7985 & 0.5674 \\
    EdenLive-5B~\cite{vast2026edenlive} & 56.8873 & 0.4319 & 15.2433 & 0.5595 & 23.1180 & 0.6537 & 0.043253 & 15.0065 & 0.5536 \\
    EdenLive-14B~\cite{vast2026edenlive} & 64.3274 & 0.4999 & \underline{17.4653} & \underline{0.6072} & 23.2588 & 0.6804 & 0.028916 & 17.1301 & 0.5913 \\
    \textbf{MemoRenderer} & \textbf{67.9704} & \underline{0.5363} & \textbf{19.0117} & \textbf{0.6987} & \textbf{23.7272} & \textbf{0.7492} & \textbf{0.009954} & \textbf{22.7539} & \textbf{0.6588} \\
    \midrule
    \multicolumn{10}{l}{\textit{(B) Geometric conditioning (50\% static + 50\% dynamic)}} \\
    \addlinespace[2pt]
    Normal + Instance & \underline{65.5149} & 0.5201 & 16.0203 & 0.5919 & 23.3852 & 0.6944 & 0.039409 & 16.4830 & 0.5991 \\
    Normal + Instance + WCM & 64.8969 & 0.5090 & 15.9570 & 0.5899 & \underline{23.4808} & 0.6899 & \underline{0.036736} & 16.4399 & 0.5994 \\
    Normal + Instance + CCM & 64.8165 & \textbf{0.5386} & \underline{16.0597} & \underline{0.6033} & 23.3916 & \underline{0.6984} & 0.039976 & \underline{16.7237} & \underline{0.6034} \\
    \textbf{Normal + Instance + WCM + CCM} & \textbf{67.9704} & \underline{0.5363} & \textbf{19.0117} & \textbf{0.6987} & \textbf{23.7272} & \textbf{0.7492} & \textbf{0.009954} & \textbf{22.7539} & \textbf{0.6588} \\
    \midrule
    \multicolumn{10}{l}{\textit{(C) Memory access strategies (50\% static + 50\% dynamic)}} \\
    \addlinespace[2pt]
    Without retrieval & 66.6989 & 0.5228 & 18.3502 & 0.6825 & 23.6954 & \textbf{0.7507} & 0.016200 & 19.9162 & 0.6337 \\
    Learned router query & 66.6857 & 0.5229 & \underline{18.6151} & 0.6923 & \underline{23.7258} & 0.7493 & 0.015651 & 20.2157 & 0.6371 \\
    Sparse-attention & \textbf{68.1288} & \underline{0.5311} & 18.4302 & 0.6812 & 23.6304 & 0.7466 & 0.014793 & 20.5991 & 0.6488 \\
    Test-time training~\cite{sun2025ttt} & 60.7117 & 0.4730 & 18.0347 & \underline{0.6926} & 23.6576 & 0.7396 & 0.013817 & 20.3528 & 0.6289 \\
    Retrieval in full history & 66.8006 & 0.5238 & 18.5993 & 0.6914 & 23.6949 & \underline{0.7501} & \underline{0.011601} & \underline{21.3756} & \underline{0.6497} \\
    \textbf{Retrieval in memory bank (Ours)} & \underline{67.9704} & \textbf{0.5363} & \textbf{19.0117} & \textbf{0.6987} & \textbf{23.7272} & 0.7492 & \textbf{0.009954} & \textbf{22.7539} & \textbf{0.6588} \\
    \bottomrule
  \end{tabular*}
\end{table*}

\paragraph{Evaluation Dataset.}
We evaluate \method{} on 27 static scenes and 16 code-driven dynamic sequences. Static evaluation uses both 81-frame prefixes and full 401-frame rollouts, while dynamic evaluation uses 960-frame rollouts. All sequences run at 16 fps and are scored on an $832\times480$ grid. CodeMemBench organizes memory evaluation around static appearance persistence, moving-instance identity, multi-view integration, and program-controlled state changes (\cref{fig:codemembench}). Viewpoint change, observation delay, repeated visits, and similar-instance distractors vary across cases. We distinguish appearance visible in the initial image from appearance first generated during rollout. When world states change, the evaluation retains historical references for attributes that should persist and accounts for the updated state of changed attributes.

\paragraph{Baselines.}
We compare against code-driven renderers AgentGarten~\cite{chi2026agentgarten}, Streaming CWM~\cite{liao2026streamingcwm}, SolarRenderer~\cite{fang2026solarrenderer}, and EdenLive-5B/14B~\cite{vast2026edenlive}, together with geometry-conditioned methods VACE~\cite{vace}, DaS~\cite{gu2025das}, LongVie~2~\cite{gao2025longvie2}, VideoFrom3D~\cite{kim2025videofrom3d}, and LTX-2.3 (3DREAL)~\cite{hacohen2026ltx,odin2026_3dreal}. The short-clip comparison additionally includes C2R~\cite{gomeznogales2026coarsetoreal}. Each renderer receives its supported condition representation exported from the same scene state and camera trajectory, with the same initial visual reference and text prompt where supported. Controlled comparisons of memory mechanisms are presented in \cref{subsec:ablation}.

\paragraph{Metrics.}
We assess four complementary dimensions. \textit{Visual quality} uses frame-level MUSIQ~\cite{ke2021musiq} and video-level DOVER~\cite{wu2023dover}. \textit{Visual alignment} uses PSNR and SSIM against reference RGB. \textit{Structural alignment} uses PSNR-D~\cite{kim2025videofrom3d}, which compares predicted depth~\cite{yang2024depthv2} with engine depth, and control-aligned DINO patch similarity (Ctrl.~DINO)~\cite{simeoni2025dinov3}, which compares generated and normal-map features at corresponding spatial locations. \textit{Memory consistency} uses M.MSE, M.PSNR, and M.DINO~\cite{oquab2024dinov2} between geometrically corresponding patches in the current output and the same rollout's generated history. These scores use common valid surface regions and correspondences fixed independently of retrieval. Metrics are averaged equally across eligible cases. Long-rollout comparisons weight the static and dynamic benchmark means equally. Further sampling, coverage, and aggregation details are in the supplementary material.

\subsection{Comparison Results}
\label{subsec:main_results}

\paragraph{Quantitative Comparison.}
\Cref{tab:renderer_comparison} and \cref{tab:long_rollout}(A) summarize the short-clip and long-rollout comparisons. Across both settings, MemoRenderer achieves state-of-the-art performance in visual alignment (PSNR, SSIM), control-aligned DINO similarity, and frame-level quality (MUSIQ), while also ranking first in DOVER on short clips and remaining competitive on long rollouts. On long rollouts, it further leads in depth alignment and across all three history-based memory consistency metrics, with the same trend holding separately on static and dynamic benchmarks (see the supplementary material). Evaluation on the static subset of surfaces first observed during rollout further confirms that this consistency extends beyond the initial conditioning frame. Overall, MemoRenderer excels in structural alignment and long-term appearance consistency.

\paragraph{Qualitative Comparison.}
\Cref{fig:comparison_short_term} illustrates qualitative comparisons across representative memory behaviors. These qualitative examples use the base renderer to isolate memory behavior. Specifically, the \textit{interior} scene evaluates \textit{static appearance persistence} under large viewpoint changes, where the cabinet layout and surface textures should remain unchanged across revisits. The \textit{kart-racing} sequence evaluates \textit{moving-instance identity} amidst nearby distractors, where both the vehicle and driver appearance should persist over time. MemoRenderer faithfully preserves these visual attributes across paired observations, whereas baseline renderers frequently alter cabinet textures or drift in kart appearance and positioning. Additional farmhouse and character comparisons are provided in the supplementary material.

\subsection{Ablation Studies}
\label{subsec:ablation}

\paragraph{Geometric Conditioning.}
\Cref{tab:long_rollout}(B) evaluates normal and instance maps with WCMs and CCMs added separately and jointly. All variants retain the same numerical geometry for engine-side retrieval; the video model receives only the listed condition maps. Initial inputs, seeds, sampling, and the memory interface are matched within each benchmark. Our full four-map configuration achieves the best results on eight of the nine reported metrics, including all visual-alignment, structure-alignment, and memory-consistency metrics. WCMs describe surfaces in a shared world frame, while instance-level CCMs provide an object-centric coordinate scale and stable correspondences under motion. Together, these geometric cues support structure preservation and consistent historical appearance. Separate static and dynamic results are reported in the supplementary material.

\begin{figure}[t]
  \centering
  \includegraphics[width=\linewidth]{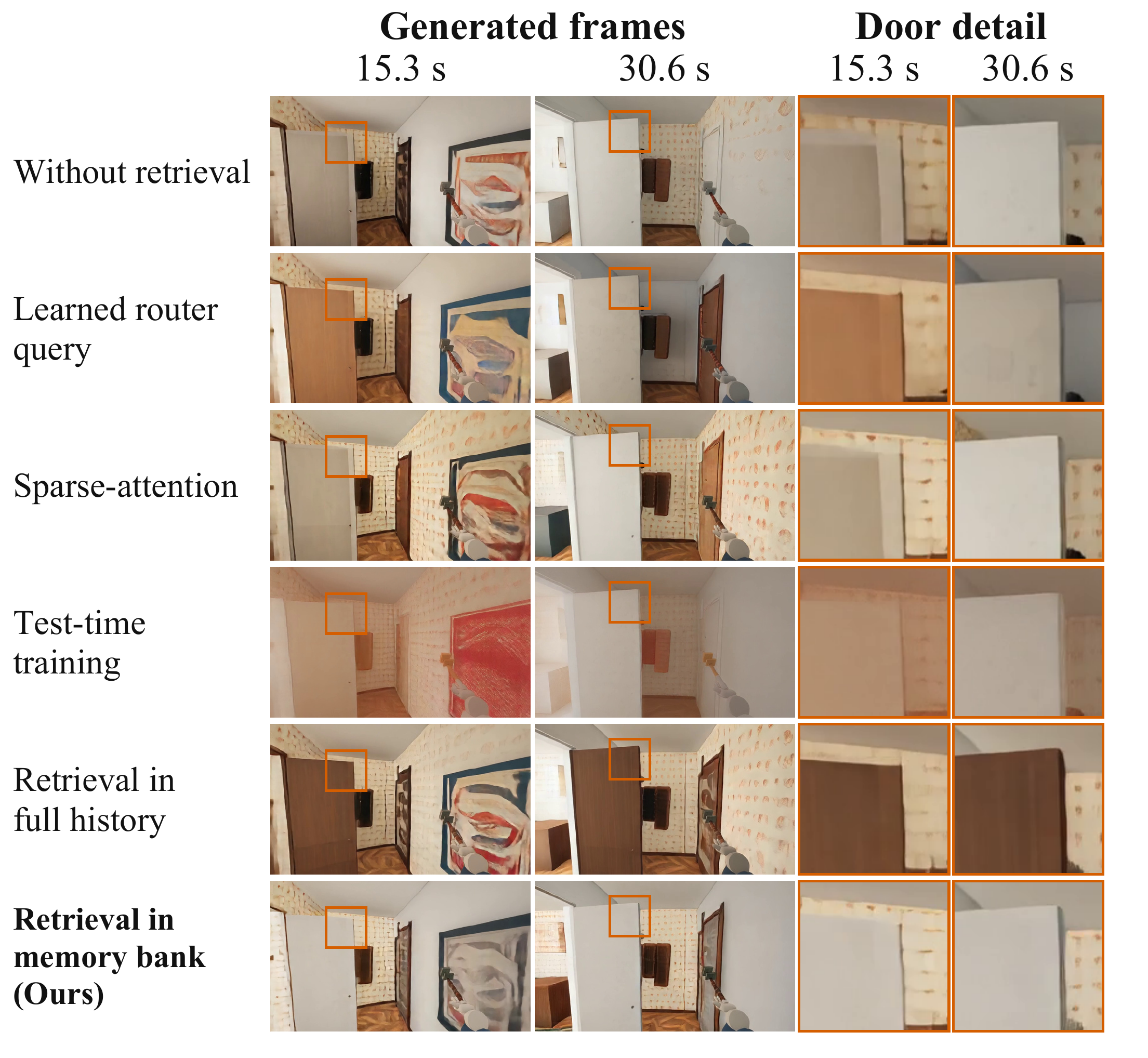}
  \caption{\textbf{Qualitative ablation on memory mechanism.} Paired observations at 15.3~s and 30.6~s show the same moving door.}
  \label{fig:memory-retrieval-ablation}
\end{figure}

\paragraph{Memory Mechanism.}
\Cref{tab:long_rollout}(C) compares six memory access strategies under a shared backbone, four-map conditioning, and a fixed attention budget.
Incorporating historical retrieval consistently improves visual alignment and memory consistency over the no-retrieval baseline. Among all evaluated methods, our instance-organized memory bank (\textit{Retrieval in memory bank}) achieves the strongest overall performance, ranking first in visual alignment (PSNR, SSIM), depth alignment, DOVER, and all three memory consistency metrics. It confirms that explicit engine-provided surface correspondences offer a more reliable signal for long-horizon consistency than learned memory heuristics. \cref{fig:memory-retrieval-ablation} illustrates how our approach better preserves previously generated door appearance under object motion.
By filtering redundant history to narrow the search scope, our compact memory bank builds clearer historical associations than exhaustive search (\textit{Retrieval in full history}), achieving stronger visual quality and memory consistency. Compared with full-history retrieval, M.MSE decreases by $14.2\%$, while M.PSNR increases by $1.38$\,dB. Alongside these gains, retrieval-only timing at frame 960 (50 queries per scene across five scenes) shows an approximately $1.96\times$ speedup over full-history search. Per-benchmark results are in the supplementary material.

\section{Conclusion and Limitations}
\label{sec:conclusion}

\method{} uses normal and instance maps to constrain structure, along with WCMs and CCMs to associate surface observations across time.
The system organizes a persistent, bounded instance-based memory bank, which is searched for key-value cache of relevant patches to preserve appearance consistency with generated history.
Experiments across static and dynamic scenes demonstrate improved retention of self-generated appearance.
Ablations favor compact, deduplicated memory over learnable alternatives and full-history retrieval.
Currently, \method{} operates at a moderate spatial resolution ($832\times480$), which can be scaled to higher resolutions with larger training budgets.
It depends on reliable geometry and tracking from the underlying engine, which can be built with advanced coding agents like Astra and Fable.
Pairing these agents with \method{} provides a promising approach to generating high-quality interactive worlds.

{\small
\bibliographystyle{ieeenat_fullname}
\bibliography{main}
}
\clearpage
\appendix
\renewcommand{\thetable}{A\arabic{table}}
\setcounter{table}{0}
\setcounter{dbltopnumber}{3}
\section{Backbone and Training Configuration}
\label{sec:supp_training}

\subsection{Backbone}
MemoRenderer uses Wan2.1-T2V-1.3B~\cite{wan2025}: 30 transformer blocks, hidden width 1,536, and 12 attention heads. An 81-frame clip at $832\times480$ has a $16\times21\times60\times104$ latent tensor (channels, time, height, width), with $1\times2\times2$ transformer patches. Latents of WCMs, normals, CCMs, and instance maps are concatenated with the noisy video latent, expanding the input projection from 16 to 80 channels. New control-channel weights are zero-initialized when constructing the conditioned backbone. The full transformer is trainable; the VAE and text encoder remain frozen.

\subsection{Training}
\paragraph{Supervised stages.}
We first run bidirectional SFT and use its EMA checkpoint (step 19,000) to initialize teacher-forcing (TF) SFT, whose EMA checkpoint (step 18,000) in turn initializes the SGF generator. Both supervised stages use 81-frame, 16-fps clips, cached video and control latents, text dropout 0.1, EMA decay 0.999, weight decay 0.01, and gradient clipping at 1.0. AdamW uses $\beta=(0.9,0.999)$ for bidirectional SFT and $(0.9,0.95)$ for TF. TF uses three-latent-frame chunks with clean ground-truth history. \Cref{tab:supp_training} summarizes the stage-local checkpoints, global batch sizes, and learning rates.

\begin{table}[b]
\centering
\caption{Selected stage-local checkpoints and learning rates. Batch sizes are global.}
\label{tab:supp_training}
\small
\setlength{\tabcolsep}{2.5pt}
\renewcommand{\arraystretch}{1.12}
\begin{tabular*}{\columnwidth}{@{\extracolsep{\fill}}lrrrr@{}}
\toprule
\textbf{Stage} & \textbf{Step} & \textbf{Batch} & \textbf{LR (G)} & \textbf{LR (fake)} \\
\midrule
Bidirectional SFT & 19,000 & 32 & $10^{-5}$ & -- \\
TF SFT & 18,000 & 32 & $2{\times}10^{-5}$ & -- \\
\midrule
SGF distillation & 1,500 & 8 & $2{\times}10^{-6}$ & $4{\times}10^{-7}$ \\
Memory adaptation & 900 & 8 & $2{\times}10^{-6}$ & $4{\times}10^{-7}$ \\
\bottomrule
\end{tabular*}
\end{table}

\paragraph{SGF distillation.}
On 81-frame clips, Self Gradient Forcing~\cite{zhuang2026sgf} uses a gradient-free rollout followed by differentiable recomputation of historical KV and current predictions from detached generated latents. Both score networks start from the bidirectional EMA (step 19,000), with the real score kept frozen. The generator uses DMD plus LPIPS~\cite{zhang2018unreasonable} with weight 1.0 and frozen AlexNet features. LPIPS samples every fourth RGB frame after native Wan VAE decoding; decoder parameters remain frozen while gradients propagate to the generated latents.

The four denoising timesteps are $(1000,937.5,833.\overline{3},625)$, with shift 5.0 and stochastic sampling. The real and fake guidance scales are 3.0 and 1.0, respectively. AdamW uses $\beta=(0,0.999)$, zero weight decay, and gradient clipping at 10.0. The generator updates once per five outer iterations and the fake score every iteration, with a generator EMA decay of 0.99 applied at each generator update.

\paragraph{Long-memory adaptation.}
The generator starts from the SGF EMA (step 1,500) and the fake score from its step-1,500 weights, both with fresh optimizers. The long-memory corpus has 3,800 training and 200 validation sequences. Episodes roll out from the sequence start to a revisit query; DMD and LPIPS supervise only the query chunk, with replay gradients reaching the retrieved historical KV. Optimization follows SGF, except that sampling is deterministic. Quantitative evaluation uses the adaptation EMA (step 900).

\paragraph{Distributed training and rollout settings.}
Training uses BF16 FSDP2, FP32 reductions, activation checkpointing, constant learning rates, random seed 42, and no gradient accumulation, running on 32 GPUs for SFT and 8 GPUs for SGF distillation and long-memory adaptation. Both adaptation and evaluation use three-latent-frame chunks, clean KV, one sink frame, a 21-latent-frame attention budget, RoPE window reindexing, and a 14,040-token retrieval cap.

\section{Memory Bank Implementation}
\label{sec:supp_memory_implementation}
\paragraph{Surface addresses and storage.}
At $832\times480$, each latent frame contains $30\times52=1{,}560$ spatial tokens. The bank stores clean, pre-RoPE K and V from all 30 transformer layers, with their source-frame and spatial indices. An entry is indexed by instance ID and a canonical cell obtained by clipping normalized CCM coordinates to $[0,1]$ and quantizing each axis into 64 bins. Instance identity remains part of the key, so identical canonical coordinates on different objects do not collide. Unsupported geometry receives a unique observation key rather than a spurious surface match.

\paragraph{Capacity control.}
We set the per-instance capacity to $M_{\mathrm{inst}}=32{,}000$ token entries and the total bank capacity to $M_{\mathrm{bank}}=256{,}000$ entries. Repeated observations are consolidated by instance and canonical-grid cell. When either capacity is exceeded, entries are pruned to retain representative views within $|\mathcal B_i[j]|\le M_{\mathrm{inst}}$ and $\sum_j|\mathcal B_i[j]|\le M_{\mathrm{bank}}$. Retention prioritizes the earliest observations, followed by spatially dispersed recent views and uniform filling. The sink and eight most recent latent frames are protected from bank pruning. Retention uses only past geometry and observation order, without RGB similarity or future query information. These storage limits are separate from the 14,040-token retrieval budget used for attention.

\paragraph{Retrieval and attention.}
Before denoising, current geometry identifies candidates from completed historical chunks. Rigid assets use historical object transforms; deforming surfaces use bind-pose triangles and barycentric coordinates to locate the same material point in historical poses. Matching verifies instance identity, canonical surface region, historical visibility, and projected footprint. The earliest valid observation still retained in the bank is prioritized for each query. Additional observations are ranked by same-instance pixel coverage, canonical sampling density, and view alignment, with geometric error, time, and source ID as tie-breakers. Round-robin selection across queries and source-ID deduplication precede uniform filling from remaining eligible retained history. Uniform fill supplies context and is not treated as a certified surface correspondence. Once evicted, historical KV entries are never fetched back from uncompressed full history.

The maximum retrieved budget is 14,040 tokens (nine latent-frame equivalents). At full occupancy, the active context comprises one sink frame, nine retrieved frame equivalents, eight recent frames, and the current three-frame chunk, totaling 21 latent frames. Before sufficient long-term history accumulates, the same total budget is filled from available causal history. Historical keys preserve spatial indices and receive temporal RoPE once when context is assembled. Only after generation does a clean forward pass commit the new chunk for subsequent queries.

\paragraph{Memory baseline implementations.}
The strategies in \cref{tab:appendix_memory_access} follow the same backbone, training pipeline, training data, and attention budget as in the main paper, differing only in the history-access mechanism.

\textit{Learned router query.} Spatially pooled clean keys from the previous generated chunk form the routing query. A shared MLP and separate query and key projections score historical blocks. The highest-scoring blocks supply their original patch KV, with learned gates applied to the selected values. The selected blocks remain fixed throughout denoising of the current chunk.

\textit{Sparse-attention.} Native attention queries and historical keys are pooled over $2\times2$ spatial blocks for relevance scoring. Top-ranked historical blocks are expanded back to their original patch KV and concatenated with the sink, recent context, and current-chunk KV for joint attention. The routing rule adds no separate learnable parameters.

\textit{Test-time training.} Each transformer layer maintains a SwiGLU fast-weight memory updated using the clean keys and values of completed chunks. Current queries read this state through a gated residual added to local attention. The state stays fixed during denoising and is updated once each chunk is finalized.

\section{CodeMemBench Composition}
\label{sec:supp_benchmark}
\begin{table}[t]
\centering
\caption{Evaluation coverage. Memory counts use geometry-only revisit eligibility, independently of renderer output quality.}
\label{tab:supp_benchmark_coverage}
\footnotesize
\setlength{\tabcolsep}{3pt}
\begin{tabular*}{\columnwidth}{@{\extracolsep{\fill}}lrrr@{}}
\toprule
Benchmark & Frames & All cases & Memory cases \\
\midrule
Static scenes & 401 & 27 & 17 \\
Code-driven sequences & 960 & 16 & 12 \\
\bottomrule
\end{tabular*}
\end{table}
\paragraph{Sequences and inputs.}
\Cref{tab:supp_benchmark_coverage} summarizes the sequence lengths along with standard and memory evaluation coverage. The static benchmark follows camera trajectories through fixed scenes, including viewpoint changes, disocclusion, and revisits. The code-driven benchmark contains 16 fixed 60-second recordings: combat, transport, a coffee kiosk, baseball, renovation, fractal exploration, flight, city observation, underwater motion, aerial combat, two racing settings, cycling, city response, character chess, and Rubik's-cube manipulation. This benchmark covers moving-object scenarios alongside stationary-environment and surface-binding diagnostics. All methods use the same source state, camera trajectory, timestamps, prompt, and initial RGB where supported. Model-specific controls are derived from the same numerical geometry. Native scoring uses $832\times480$ at 16 fps; temporal padding needed by an inference interface is excluded from evaluation.

\paragraph{Behavior coverage.}
The four benchmark behaviors are complementary rather than disjoint case labels. Static persistence tests appearance after camera revisits; moving-instance identity tests surface binding under object motion and similar-instance distractors; multi-view integration tests surfaces absent from the initial view; program-controlled changes distinguish persistent attributes from legitimate state changes. A single sequence can exhibit several behaviors. Numerical support and instance/canonical correspondence determine which pixels can be paired; unsupported effects are not assumed to have valid correspondences.

\paragraph{Fixed evaluation sets.}
All 27 static and 16 code-driven cases contribute to standard metrics. The memory subset is fixed before comparing outputs: a case needs at least eight reliable patches over at least two query frames under the shared revisit protocol. The primary static and code-driven memory sets contain 6,673 and 18,303 patches over 441 and 851 case--query pairs, respectively. The remaining cases are still included in the standard evaluation. Per-benchmark result and ablation tables are provided in \cref{tab:appendix_per_benchmark,tab:appendix_memory_access,tab:appendix_conditioning}.

\begin{table*}[t]
\centering
  \caption{Per-benchmark long-rollout comparisons.}
  \label{tab:appendix_per_benchmark}
  \footnotesize
  \setlength{\tabcolsep}{3pt}
  \begin{tabular*}{\textwidth}{@{\extracolsep{\fill}}lrrrrrrrrr@{}}
    \toprule
    \multirow{2}{*}{Method / setting} & \multicolumn{2}{c}{Visual Quality} & \multicolumn{2}{c}{Visual Alignment} & \multicolumn{2}{c}{Structure Alignment} & \multicolumn{3}{c}{Memory Consistency} \\
    \cmidrule(lr){2-3}\cmidrule(lr){4-5}\cmidrule(lr){6-7}\cmidrule(lr){8-10}
      & MUSIQ$\uparrow$ & DOVER$\uparrow$ & PSNR$\uparrow$ & SSIM$\uparrow$ & PSNR-D$\uparrow$ & Ctrl. DINO$\uparrow$ & M.MSE$\downarrow$ & M.PSNR$\uparrow$ & M.DINO$\uparrow$ \\
    \midrule
    \multicolumn{10}{l}{\textit{(A) Static-scene comparison (401 frames)}} \\
    \addlinespace[2pt]
    VACE~\cite{vace} & 63.6781 & 0.5140 & 10.9536 & 0.3778 & \underline{19.9865} & 0.6568 & 0.052415 & 14.4524 & 0.5404 \\
    DaS~\cite{gu2025das} & 48.1659 & 0.4159 & 12.3755 & 0.4472 & 17.6575 & 0.4815 & 0.081190 & 12.0286 & 0.3389 \\
    LongVie 2~\cite{gao2025longvie2} & 57.2025 & 0.5305 & 15.0536 & 0.5302 & 19.8551 & 0.6843 & 0.053849 & 14.9465 & 0.5629 \\
    VideoFrom3D~\cite{kim2025videofrom3d} & 59.2318 & \textbf{0.5825} & 11.7765 & 0.4678 & 18.9153 & 0.5842 & 0.075027 & 13.0487 & 0.5095 \\
    LTX-2.3 (3DREAL)~\cite{odin2026_3dreal} & 52.0871 & 0.4312 & 17.3949 & 0.6079 & 19.2997 & 0.6467 & 0.041993 & 14.5216 & 0.5050 \\
    AgentGarten~\cite{chi2026agentgarten} & \underline{65.4872} & 0.5408 & 17.7797 & 0.6013 & 19.6410 & 0.7064 & \underline{0.020948} & \underline{18.1380} & \underline{0.6453} \\
    Streaming CWM~\cite{liao2026streamingcwm} & 58.9097 & 0.4673 & 15.1602 & 0.5170 & 19.2047 & 0.6361 & 0.046462 & 14.1237 & 0.5322 \\
    SolarRenderer~\cite{fang2026solarrenderer} & 61.1457 & 0.5040 & 16.4833 & 0.5939 & 19.7941 & \underline{0.7247} & 0.049202 & 14.3935 & 0.5942 \\
    EdenLive-5B~\cite{vast2026edenlive} & 56.1459 & 0.4877 & 16.1532 & 0.5677 & 19.5953 & 0.6772 & 0.046966 & 14.5516 & 0.5308 \\
    EdenLive-14B~\cite{vast2026edenlive} & 62.8851 & 0.5380 & \underline{18.6476} & \underline{0.6225} & 19.6673 & 0.7116 & 0.027752 & 17.0951 & 0.6190 \\
    \textbf{MemoRenderer} & \textbf{66.7746} & \underline{0.5781} & \textbf{20.8591} & \textbf{0.7330} & \textbf{20.1345} & \textbf{0.7579} & \textbf{0.009866} & \textbf{23.1477} & \textbf{0.6829} \\
    \midrule
    \multicolumn{10}{l}{\textit{(B) Dynamic-scene comparison (960 frames)}} \\
    \addlinespace[2pt]
    VACE~\cite{vace} & 59.7857 & 0.4137 & 8.5389 & 0.3425 & 26.7190 & 0.6436 & 0.077728 & 13.0172 & 0.5366 \\
    DaS~\cite{gu2025das} & 51.9866 & 0.3624 & 10.7332 & 0.4097 & 25.3371 & 0.4429 & 0.072424 & 13.2239 & 0.4071 \\
    LongVie 2~\cite{gao2025longvie2} & 62.8027 & 0.4326 & 11.8858 & 0.4299 & 26.2097 & 0.5723 & 0.069710 & 12.8523 & 0.4818 \\
    VideoFrom3D~\cite{kim2025videofrom3d} & 62.4470 & \textbf{0.5113} & 12.8985 & 0.4898 & 26.3791 & 0.5797 & 0.040339 & 14.8159 & 0.5078 \\
    LTX-2.3 (3DREAL)~\cite{odin2026_3dreal} & 57.4298 & 0.4276 & 12.4038 & 0.4856 & 26.0243 & 0.5270 & 0.048598 & 13.4072 & 0.4533 \\
    AgentGarten~\cite{chi2026agentgarten} & \underline{67.8373} & 0.4414 & 13.9876 & 0.5283 & 26.4825 & 0.6343 & \underline{0.023040} & \underline{18.0579} & 0.5741 \\
    Streaming CWM~\cite{liao2026streamingcwm} & 62.9614 & 0.4107 & 13.1962 & 0.4342 & 26.6062 & 0.5874 & 0.023548 & 16.8113 & 0.4903 \\
    SolarRenderer~\cite{fang2026solarrenderer} & 63.6430 & 0.4125 & 12.2732 & 0.5220 & 26.4913 & \underline{0.6627} & 0.054256 & 13.2034 & 0.5406 \\
    EdenLive-5B~\cite{vast2026edenlive} & 57.6287 & 0.3760 & 14.3334 & 0.5513 & 26.6406 & 0.6302 & 0.039539 & 15.4613 & \underline{0.5763} \\
    EdenLive-14B~\cite{vast2026edenlive} & 65.7697 & 0.4618 & \underline{16.2830} & \underline{0.5919} & \underline{26.8503} & 0.6492 & 0.030079 & 17.1650 & 0.5636 \\
    \textbf{MemoRenderer} & \textbf{69.1663} & \underline{0.4945} & \textbf{17.1644} & \textbf{0.6645} & \textbf{27.3199} & \textbf{0.7405} & \textbf{0.010043} & \textbf{22.3602} & \textbf{0.6347} \\
    \bottomrule
  \end{tabular*}
\end{table*}

\begin{table*}[t]
\centering
  \caption{Memory access ablation by benchmark.}
  \label{tab:appendix_memory_access}
  \footnotesize
  \setlength{\tabcolsep}{3pt}
  \begin{tabular*}{\textwidth}{@{\extracolsep{\fill}}lrrrrrrrrr@{}}
    \toprule
    \multirow{2}{*}{Method / setting} & \multicolumn{2}{c}{Visual Quality} & \multicolumn{2}{c}{Visual Alignment} & \multicolumn{2}{c}{Structure Alignment} & \multicolumn{3}{c}{Memory Consistency} \\
    \cmidrule(lr){2-3}\cmidrule(lr){4-5}\cmidrule(lr){6-7}\cmidrule(lr){8-10}
      & MUSIQ$\uparrow$ & DOVER$\uparrow$ & PSNR$\uparrow$ & SSIM$\uparrow$ & PSNR-D$\uparrow$ & Ctrl. DINO$\uparrow$ & M.MSE$\downarrow$ & M.PSNR$\uparrow$ & M.DINO$\uparrow$ \\
    \midrule
    \multicolumn{10}{l}{\textit{(A) Static-scene memory access (401 frames)}} \\
    \addlinespace[2pt]
    Without retrieval & 65.6747 & 0.5676 & 20.3363 & 0.7166 & \textbf{20.1524} & 0.7572 & 0.013726 & 21.0322 & 0.6621 \\
    Learned router query & 65.5410 & 0.5657 & 20.5055 & 0.7233 & \underline{20.1420} & 0.7564 & 0.012194 & 22.0355 & 0.6724 \\
    Sparse-attention & \textbf{67.0146} & \underline{0.5739} & 20.4995 & 0.7231 & 20.1340 & \textbf{0.7582} & 0.010226 & 22.3716 & \underline{0.6780} \\
    Test-time training~\cite{sun2025ttt} & 57.1766 & 0.4939 & 19.5078 & 0.7187 & 20.0779 & 0.7437 & 0.013378 & 20.7141 & 0.6542 \\
    Retrieval in full history & 65.6842 & 0.5687 & \underline{20.5227} & \underline{0.7242} & 20.1299 & 0.7560 & \textbf{0.008290} & \textbf{23.3020} & 0.6764 \\
    \textbf{Retrieval in memory bank (Ours)} & \underline{66.7746} & \textbf{0.5781} & \textbf{20.8591} & \textbf{0.7330} & 20.1345 & \underline{0.7579} & \underline{0.009866} & \underline{23.1477} & \textbf{0.6829} \\
    \midrule
    \multicolumn{10}{l}{\textit{(B) Dynamic-scene memory access (960 frames)}} \\
    \addlinespace[2pt]
    Without retrieval & 67.7230 & 0.4780 & 16.3641 & 0.6484 & 27.2384 & \underline{0.7441} & 0.018673 & 18.8002 & 0.6053 \\
    Learned router query & 67.8303 & 0.4800 & \underline{16.7247} & 0.6613 & \underline{27.3096} & 0.7422 & 0.019107 & 18.3958 & 0.6018 \\
    Sparse-attention & \textbf{69.2429} & \underline{0.4883} & 16.3608 & 0.6392 & 27.1268 & 0.7349 & 0.019360 & 18.8265 & 0.6196 \\
    Test-time training~\cite{sun2025ttt} & 64.2468 & 0.4521 & 16.5616 & \textbf{0.6665} & 27.2373 & 0.7354 & \underline{0.014255} & \underline{19.9914} & 0.6036 \\
    Retrieval in full history & 67.9170 & 0.4788 & 16.6758 & 0.6586 & 27.2600 & \textbf{0.7443} & 0.014912 & 19.4491 & \underline{0.6230} \\
    \textbf{Retrieval in memory bank (Ours)} & \underline{69.1663} & \textbf{0.4945} & \textbf{17.1644} & \underline{0.6645} & \textbf{27.3199} & 0.7405 & \textbf{0.010043} & \textbf{22.3602} & \textbf{0.6347} \\
    \bottomrule
  \end{tabular*}
\end{table*}

\begin{table*}[t]
\centering
  \caption{Geometric conditioning ablation by benchmark with fixed engine geometry and memory interface.}
  \label{tab:appendix_conditioning}
  \label{tab:supp_ccm_ablation}
  \footnotesize
  \setlength{\tabcolsep}{3pt}
  \begin{tabular*}{\textwidth}{@{\extracolsep{\fill}}lrrrrrrrrr@{}}
    \toprule
    \multirow{2}{*}{Method / setting} & \multicolumn{2}{c}{Visual Quality} & \multicolumn{2}{c}{Visual Alignment} & \multicolumn{2}{c}{Structure Alignment} & \multicolumn{3}{c}{Memory Consistency} \\
    \cmidrule(lr){2-3}\cmidrule(lr){4-5}\cmidrule(lr){6-7}\cmidrule(lr){8-10}
      & MUSIQ$\uparrow$ & DOVER$\uparrow$ & PSNR$\uparrow$ & SSIM$\uparrow$ & PSNR-D$\uparrow$ & Ctrl. DINO$\uparrow$ & M.MSE$\downarrow$ & M.PSNR$\uparrow$ & M.DINO$\uparrow$ \\
    \midrule
    \multicolumn{10}{l}{\textit{(A) Static-scene conditioning ablation (401 frames)}} \\
    \addlinespace[2pt]
    Normal + Instance & \underline{64.4188} & 0.5609 & 17.8991 & 0.6420 & \underline{20.0284} & 0.7279 & 0.024256 & 18.1954 & 0.6372 \\
    Normal + Instance + WCM & 63.5735 & 0.5554 & 17.6636 & 0.6372 & 20.0036 & 0.7232 & 0.023967 & 18.1397 & 0.6368 \\
    Normal + Instance + CCM & 63.8158 & \underline{0.5679} & \underline{18.0983} & \underline{0.6557} & 19.9894 & \underline{0.7345} & \underline{0.019863} & \underline{18.6951} & \underline{0.6388} \\
    \textbf{Normal + Instance + WCM + CCM} & \textbf{66.7746} & \textbf{0.5781} & \textbf{20.8591} & \textbf{0.7330} & \textbf{20.1345} & \textbf{0.7579} & \textbf{0.009866} & \textbf{23.1477} & \textbf{0.6829} \\
    \midrule
    \multicolumn{10}{l}{\textit{(B) Dynamic-scene conditioning ablation (960 frames)}} \\
    \addlinespace[2pt]
    Normal + Instance & \underline{66.6109} & 0.4793 & 14.1415 & 0.5418 & 26.7420 & 0.6609 & 0.054562 & \underline{14.7705} & 0.5610 \\
    Normal + Instance + WCM & 66.2202 & 0.4625 & \underline{14.2504} & 0.5426 & \underline{26.9580} & 0.6566 & \underline{0.049505} & 14.7401 & 0.5619 \\
    Normal + Instance + CCM & 65.8171 & \textbf{0.5093} & 14.0211 & \underline{0.5509} & 26.7939 & \underline{0.6622} & 0.060090 & 14.7522 & \underline{0.5681} \\
    \textbf{Normal + Instance + WCM + CCM} & \textbf{69.1663} & \underline{0.4945} & \textbf{17.1644} & \textbf{0.6645} & \textbf{27.3199} & \textbf{0.7405} & \textbf{0.010043} & \textbf{22.3602} & \textbf{0.6347} \\
    \bottomrule
  \end{tabular*}
\end{table*}

\paragraph{Training and evaluation units.}
The short training corpus contains 19,000 training and 1,000 validation clips, while long-memory adaptation uses 3,800 training and 200 validation sequences. Both follow the source-session split, and long sequences retain their corresponding session assignment rather than being split independently by clip. Evaluation follows the separately fixed 27-scene static and 16-sequence dynamic manifests.

\section{Additional Conditioning Comparison}
\label{sec:supp_ccm_ablation}
\Cref{tab:appendix_conditioning} reports the static and dynamic conditioning ablations separately, complementing the equally weighted results in \cref{tab:long_rollout}(B). Our full four-map configuration ranks first on all nine static-scene metrics and eight of nine dynamic-scene metrics. It leads all three memory-consistency metrics on both benchmarks, reaching M.PSNR of 23.1477\,dB on static scenes and 22.3602\,dB on dynamic scenes. These results highlight the complementary roles of WCMs and CCMs in structural alignment and historical appearance consistency.

\begin{table*}[t]
\centering
\caption{Static-scene memory consistency by initial visibility (401 frames). Both strata use fixed geometry pairs and each method's own generated history. Initial-frame surfaces cover 14 cases, 310 case--query pairs, and 3,325 patches; newly observed surfaces cover 12 cases, 251 case--query pairs, and 2,488 patches.}
\label{tab:supp_surface_memory}
\footnotesize
\setlength{\tabcolsep}{4pt}
\begin{tabular*}{\textwidth}{@{\extracolsep{\fill}}lrrrrrr@{}}
\toprule
\multirow{2}{*}{Method} & \multicolumn{3}{c}{Visible in the initial frame} & \multicolumn{3}{c}{First observed during rollout} \\
\cmidrule(lr){2-4}\cmidrule(lr){5-7}
& M.MSE$\downarrow$ & M.PSNR$\uparrow$ & M.DINO$\uparrow$ & M.MSE$\downarrow$ & M.PSNR$\uparrow$ & M.DINO$\uparrow$ \\
\midrule
VACE & 0.055272 & 14.1220 & 0.5674 & 0.054632 & 14.6128 & 0.5612 \\
DaS & 0.085414 & 11.7389 & 0.3254 & 0.072017 & 12.0313 & 0.3814 \\
LongVie 2 & 0.063803 & 14.3475 & 0.5559 & 0.056776 & 14.2830 & 0.5805 \\
VideoFrom3D & 0.077015 & 13.0409 & 0.5291 & 0.054984 & 13.8442 & 0.5197 \\
LTX-2.3 (3DREAL) & 0.041354 & 14.4996 & 0.5186 & 0.035108 & 15.2181 & 0.5062 \\
AgentGarten & \underline{0.016842} & \underline{19.0843} & \underline{0.6731} & \underline{0.028275} & \underline{16.8353} & \underline{0.6323} \\
Streaming CWM & 0.045747 & 14.1640 & 0.5170 & 0.044771 & 14.3283 & 0.5180 \\
SolarRenderer & 0.042003 & 15.5636 & 0.6261 & 0.045406 & 14.1199 & 0.5940 \\
EdenLive-5B & 0.053360 & 14.0206 & 0.5403 & 0.065623 & 13.0608 & 0.4828 \\
EdenLive-14B & 0.023527 & 17.7967 & 0.6597 & 0.035984 & 15.8336 & 0.5674 \\
\textbf{MemoRenderer} & \textbf{0.006190} & \textbf{24.9345} & \textbf{0.7138} & \textbf{0.012580} & \textbf{22.0872} & \textbf{0.6645} \\
\bottomrule
\end{tabular*}
\end{table*}

\section{Evaluation Details}
\label{sec:evaluation_protocol}

\paragraph{History-based memory.}
The reported memory metrics compare generated RGB with earlier generated observations of the same surface, using frozen geometry-derived correspondences rather than ground-truth appearance or the renderer's retrieval choices. Geometry pairing uses stride-five frames. We score native $32\times32$ patches with at least 90\% valid pixels and at most one-pixel reprojection roundtrip error. The primary revisit pairs have a reference age of at least five seconds and out-of-view evidence spanning at least two seconds. References use the earliest sufficiently clear observation. A case is eligible if it contains at least eight reliable patches across at least two query frames.

Within each benchmark, let $e_c$ be the valid-pixel RGB MSE for case $c$, after warping historical RGB into query coordinates and averaging equally over patches within each query and then over query frames. RGB lies in $[0,1]$, and $\mathcal C$ is the eligible case set. The reported scores are
\begin{equation}
\begin{aligned}
\mathrm{M.MSE}&=\frac{1}{|\mathcal C|}\sum_{c\in\mathcal C}e_c,\\
\mathrm{M.PSNR}&=\frac{1}{|\mathcal C|}\sum_{c\in\mathcal C}-10\log_{10}\!\bigl(\max(e_c,10^{-12})\bigr).
\end{aligned}
\end{equation}
M.DINO~\cite{oquab2024dinov2} uses final-layer DINOv2-B/14 patch features from native-resolution images, replicate-padded to multiples of 14. Current and historical features are bilinearly sampled at corresponding pixels; cosine similarity is averaged over the same valid support using the same per-query and per-case weighting.

\paragraph{Structural alignment.}
PSNR-D~\cite{kim2025videofrom3d} uses Depth Anything V2 Large~\cite{yang2024depthv2}. Reference depth is float32 camera-axis Z in scene units, normalized by a fixed per-case range. Predicted depth is scale-shift fitted over each evaluation window and clipped to $[0,1]$ before scoring on pixels with valid depth support. Ctrl.~DINO~\cite{simeoni2025dinov3} compares generated RGB with the fixed raw-normal reference video using DINOv3 ViT-L/16 final-layer patch tokens. ImageNet normalization is applied at $832\times480$ without resizing or padding. CLS/register tokens are removed; FP32 cosine similarity is averaged over the $30\times52$ grid and all frames.

\paragraph{Sampling and aggregation.}
PSNR, SSIM, MUSIQ~\cite{ke2021musiq}, and depth metrics use stride-five frames on the common $832\times480$ inputs; MUSIQ and depth models retain their fixed official preprocessing. DOVER~\cite{wu2023dover} retains its official view sampling and score fusion. Memory uses the fixed 17 static and 12 dynamic eligible cases. The main comparison averages the static and dynamic benchmark means with equal weight, after per-case metric aggregation.

\paragraph{C2R.}
We include C2R~\cite{gomeznogales2026coarsetoreal} only in the 81-frame comparison because its released inference entry point provides no first-frame or reference-frame input for continuation.

\section{Memory Beyond the Initial Frame}
\label{sec:supp_surface_memory}
\Cref{tab:supp_surface_memory} groups static-scene revisit patches by initial-frame visibility. Newly observed surfaces must be confirmed absent at frame zero by scene geometry and clearly observed at a later frame. Ambiguous initial visibility is excluded from these strata but retained in all-surface evaluation where eligible. A case may contribute to both groups, with eligibility applied separately.

Scores reuse the original generated videos and frozen correspondences, comparing each method with its own earlier output. Queries are weighted equally within each case, and cases are weighted equally across each stratum. MemoRenderer leads all three memory metrics in both strata. Crucially, on newly observed surfaces, it reaches 22.09\,dB M.PSNR (+5.25\,dB over AgentGarten) and reduces M.MSE by 55.5\%, confirming that its advantage extends beyond the initial anchor to appearance generated during rollout.

\begin{table}[b]
\centering
\caption{Runtime throughput after warmup on NVIDIA H100 GPUs.}
\label{tab:supp_fps_edenstyle}
\footnotesize
\setlength{\tabcolsep}{2.5pt}
\renewcommand{\arraystretch}{1.12}
\begin{tabular*}{\columnwidth}{@{\extracolsep{\fill}}llrrr@{}}
\toprule
\textbf{Method} & \textbf{Resolution} & \shortstack{\textbf{H100}\\\textbf{GPUs}} & \shortstack{\textbf{Latent}\\\textbf{/ chunk}} & \shortstack{\textbf{Throughput}\\\textbf{(FPS)}} \\
\midrule
EdenLive 5B & $832\times480$ & 1 & 3 & 56.45 \\
EdenLive 5B & $1280\times704$ & 1 & 3 & 23.42 \\
EdenLive 14B & $832\times480$ & 8 & 3 & 16.03 \\
AgentGarten & $832\times480$ & 1 & 4 & 31.03 \\
SolarRenderer & $1344\times768$ & 8 & 5 & 1.34 \\
StreamCWM & $1344\times768$ & $7+1$ & 5 & 1.95 \\
\midrule
\textbf{MemoRenderer} & $832\times480$ & 1 & 3 & \textbf{25.95} \\
\bottomrule
\end{tabular*}
\end{table}

\section{Inference Efficiency}
\label{sec:supp_efficiency}
\paragraph{Measurement protocol.}
\Cref{tab:supp_fps_edenstyle} summarizes the measured throughput, input resolutions, GPU counts, and chunk sizes. Following EdenLive~\cite{vast2026edenlive}, we measure native-resolution throughput on H100 GPUs using three matched 30-second sequences and seed 20261028. Each case has one full warmup and three timed requests, from model-side condition processing to CPU RGB completion. We report $\mathrm{FPS}=N/T$ over all nine timed requests, where $N$ excludes the initial frame and padded outputs while padding computation remains timed. Initial loading, offline input preparation, file encoding, and networking are excluded. Static text-only embeddings are reused after warmup; visual conditioning is recomputed.

\paragraph{Implementation.}
For throughput measurement, MemoRenderer uses a BF16 DiT, TinyVAE encoding and decoding, FlashAttention-4 (FA4), and KV-cache optimizations, whereas quality evaluation retains the native Wan VAE. AgentGarten uses the Wan encoder with a TAEW22Super decoder, EdenLive uses LightTAE, and SolarRenderer shards contiguous DiT layers and their KV caches across eight GPUs.

\section{Additional Qualitative Results}
\label{sec:appendix_qualitative}
\label{sec:supp_qualitative}
\begin{figure*}[t]
  \centering
  \includegraphics[width=\textwidth]{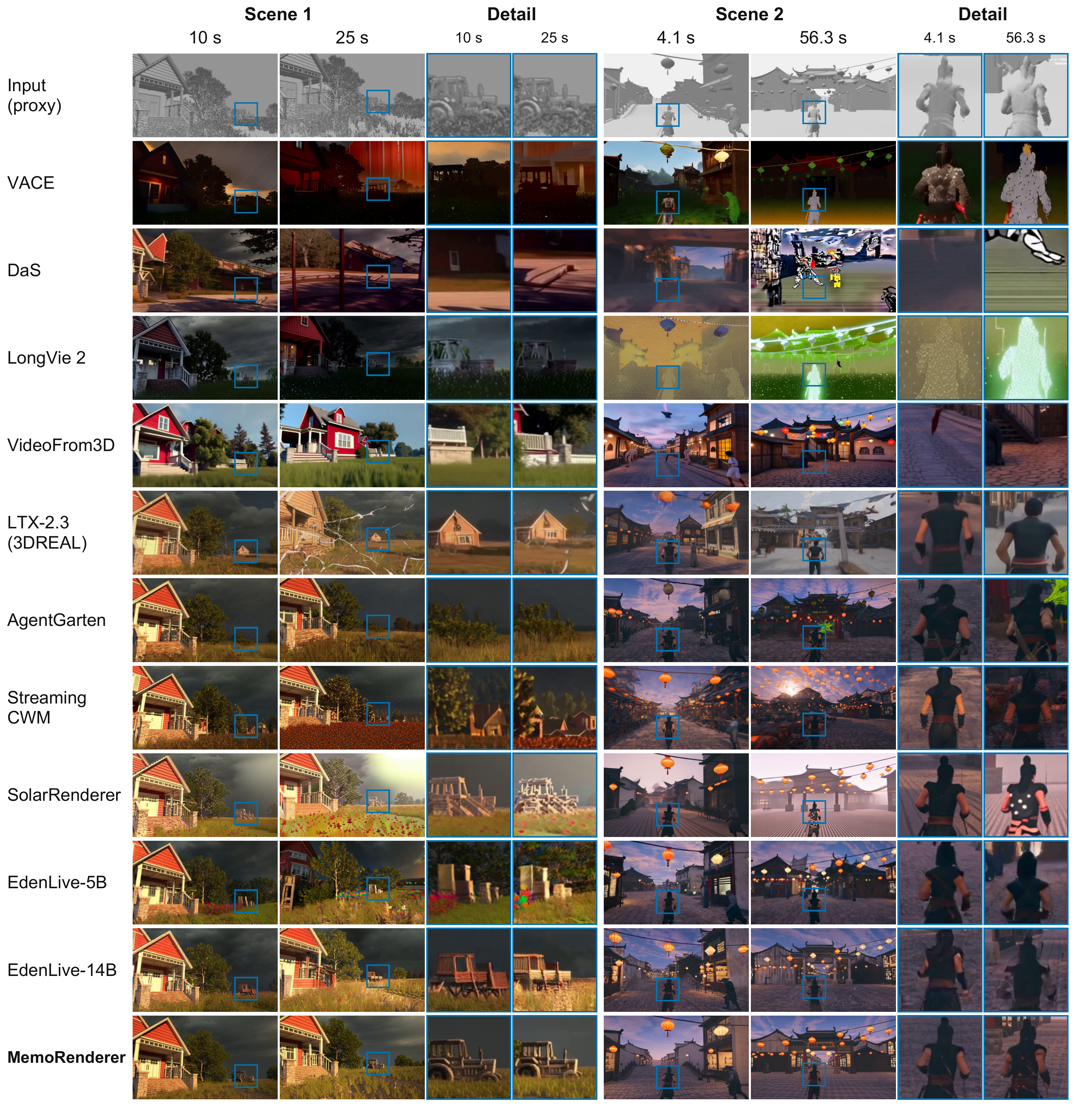}
  \caption{Additional static and dynamic comparisons. Left: farmhouse observations at 10\,s and 25\,s, with crops of the tractor region. Right: town-character observations at 4.1\,s and 56.3\,s, with crops of the character region. Each row shows one renderer under the same scene inputs; the top row shows the geometry proxy. Blue boxes locate the corresponding crops.}
  \label{fig:comparison_extra_cases}
  \label{fig:comparison_extra_character}
\end{figure*}

\Cref{fig:comparison_extra_cases} extends the main paper's base-renderer comparisons to a static farmhouse and a moving town character. Each renderer is shown at two observations with corresponding local crops. In the farmhouse scene, the tractor region reveals changes in object shape and appearance across viewpoints. In the town scene, the character crops expose changes in clothing and silhouette over a longer interval. MemoRenderer preserves recognizable tractor and character appearance across these views, whereas several baselines alter the highlighted object or deviate from the proxy geometry. These examples highlight combined structural fidelity and appearance consistency in end-to-end rollouts.

\end{document}